\pdfoutput=1

\documentclass[11pt]{article}

\usepackage{EMNLP2023}
\usepackage{graphicx}
\usepackage{multirow}
\usepackage{multicol}
\usepackage{times}
\usepackage{latexsym}
\usepackage{comment}
\usepackage[T1]{fontenc}

\usepackage[utf8]{inputenc}
\usepackage{url}
\usepackage{caption}
\usepackage{microtype}

\usepackage{inconsolata}
\usepackage{amsmath}

\title{Long-context Reference-based MT Quality Estimation }

 \author{Sami Ul Haq, Chinonso Cynthia Osuji, Sheila Castilho, Brian Davis\\
         ADAPT Centre, Dublin City University, Dublin, Ireland\\
         {\tt \{sami.haq, chinonso.osuji, sheila.castilho, brian.davis\}@adaptcentre.ie}
         }

\begin{document}
\maketitle
\begin{abstract}

In this paper, we present our submission to the Tenth Conference on Machine Translation (WMT25) Shared Task on Automated Translation Quality Evaluation. Our systems are built upon the COMET framework and trained to predict segment-level Error Span Annotation (ESA) scores using augmented long-context data. To construct long-context training data, we concatenate in-domain, human-annotated sentences and compute a weighted average of their scores. We integrate multiple human judgment datasets (MQM, SQM, and DA) by normalising their scales and train multilingual regression models to predict quality scores from the source, hypothesis, and reference translations. Experimental results show that incorporating long-context information improves correlations with human judgments compared to models trained only on short segments.

\end{abstract}

\section{Introduction}
The automatic evaluation of machine translation (MT) is a crucial component of MT research and development. While expert-based human evaluation remains the gold standard, automatic evaluation offers fast and scalable judgments, enabling rapid feedback for optimizing model parameters. Traditionally, automatic MT evaluation metrics have relied on basic lexical-level features, such as counting matching n-grams between the MT hypothesis and the reference translation. Metrics such as BLEU \citep{papineni-etal-2002-bleu}, METEOR \citep{banerjee2005meteor}, and ChrF \citep{popovic2015chrf} remain popular due to their lightweight design and computational efficiency \citep{marie2021scientific}. More recently, neural metrics (either trained on human annotations or based on pre-trained language models) have demonstrated superior capability in comparing and assessing MT quality, often outperforming traditional lexical-based metrics \citep{freitag_results_2022}. These neural approaches leverage large-scale multilingual data during training and achieve strong performance even when translations diverge lexically from the reference.

This paper presents DCU\_ADAPT's submission to the WMT25\footnote{\url{https://www2.statmt.org/wmt25/mteval-subtask.html}} MT Evaluation Shared Task. The primary focus of this year’s task is on systems capable of evaluating translation quality in context, where the context spans entire documents or multiple consecutive segments. We participated in the segment-level quality score prediction track for English–Czech, English–Russian, English–Japanese, and English–Chinese, employing models based on the COMET framework \citep{rei-etal-2020-comet}.

In our contribution to the shared task, we explore methods for leveraging synthetic data alongside the capabilities of pre-trained and cross-lingual models to predict MT quality estimates for long-sequence or multi-sentence units. Human judgments of MT quality are typically available as short segment-level scores, such as DA \citep{graham_can_2017}, MQM \citep{lommel2014multidimensional}, and SQM \cite{barrault2019findings}. Recent pre-trained models support larger context windows and can handle long-sequence inputs, improving discourse-level resolution \citep{dai2019transformer}—albeit at the cost of increased memory and computational requirements. However, most existing automatic evaluation metrics (AEMs) predict scores at the sentence level, and those designed for document-level evaluation often perform only shallow context integration during inference \citep{vernikos2022embarrassingly}. We propose a data augmentation strategy to train multilingual models on long-context annotated data, enabling them to better exploit broader context and reduce inconsistencies caused by sentence-level ambiguity.

The exploration of context in MT is a well-established topic and, in recent years, has become a focal point, driven by the need to incorporate context into both MT systems and their evaluation methodologies \citep{bawden2017evaluating, castilho-etal-2020-context, maruf2021survey, castilho2023online, castilho_survey_2024}. There is now broad consensus on the value of document-level evaluation. Since 2019, WMT has conducted human evaluations at the document level, providing evaluators with access to context even when collecting segment-level ratings \citep{akhbardeh_findings_2021, kocmi_findings_2022, kocmi_findings_2023, kocmi2024findings}. Research indicates that the appropriate context span is critical for reliable MT evaluation, with \citet{castilho-etal-2020-context} showing that incorporating relevant context spans can yield more accurate assessments of translation quality, thereby improving the evaluation process. Several techniques have been proposed to extend evaluation to the document level or to incorporate multi-sentence context into automatic evaluation metrics \citep{jiang2021blonde, vernikos2022embarrassingly, rei-etal-2022-comet, kocmi-etal-2022-ms, raunak2024slidereferencefreeevaluationmachine}.

For this shared task submission, we use the Estimator model from the COMET framework \citep{rei-etal-2022-comet}, which learns MT quality from human evaluation data such as MQM and DA. To create long-context training data, we combine multiple annotations using a weighted average alongside the original annotations. Our experiments show promising progress toward improved correlation in multi-sentence-level MT quality estimation. Fine-tuning multilingual embedding models demonstrates that it is possible to achieve high correlations with human judgments when evaluating long segments, rather than relying solely on sentence-level score predictions.

We release the data and code produced during this research.

\section{Corpora}
\label{corpora}
We used the human annotations from previous WMT shared tasks \citep{kocmi_findings_2022, kocmi_findings_2023, kocmi2024findings} for training our models, which includes human annotations from MQM, DA, and SQM. We train and evaluate our models for English (en) to Czech (cs), Japanese (ja), Chinese (zh), and Russian (ru) language pairs.
MQM scores are derived from error annotations and can range from $-\infty$ to 100. Since our goal is to predict ESA scores \citep{kocmi2024error}, which range between 0 and 100, we normalise (using Equation \ref{eq:minmax}, where $x$ is original and $x'$ is normalised score) and rescale the scores to the [0, 1] interval for training models on the combined dataset.

\begin{equation}
x' = \frac{x - \min(x)}{\max(x) - \min(x)}
\label{eq:minmax}
\end{equation}

We create augmented training data for long-span MT quality estimation by taking a weighted average of segment-level scores. During augmentation, we concatenate multiple samples (i.e., 2, 3, 4, and 5 segments) to form long-span texts. To construct the long-span MT evaluation dataset, adjacent short segments are concatenated, and a document-level quality score is computed as a length-weighted average of their original scores. The weighting is based on the total number of characters in the source and machine-translated texts, as formalized in Equation \ref{eq_long-span}.

Let $s_1$, $s_2$ are short segments (e.g., source-translation pairs), and $\text{raw}_1$, $\text{raw}_2$ human evaluation scores for these segments. $C_1$ and $C_2$ are the total character count of each segment e.g., $C_i = len(s_i)$. 

Then the document-level score ($\text{raw}_{\text{doc}}$) is calculated as:

\begin{equation}
\text{raw}_{\text{doc}} = \frac{C_1 \cdot \text{raw}_1 + C_2 \cdot \text{raw}_2}{C_1 + C_2}
\label{eq_long-span}
\end{equation}

This equation\footnote{We adapted the implementation of data augmentation from the Huggingface repository: \url{https://huggingface.co/datasets/ymoslem/wmt-da-human-evaluation-long-context}} computes a weighted average of two segment-level scores, where the weight is determined by the combined character count of the source and MT for each segment. Longer segments contribute more to the final score, reflecting their higher informational content. 

We then create training, test, and validation sets by randomly sampling segments from the training data. The statistics of the final training, validation, and test sets are shown in Table~\ref{data_stat}.

\begin{table}[ht]
\small
\centering
\begin{tabular}{llrrrr}
\hline
\textbf{Type} & \textbf{Split} & \multicolumn{4}{c}{\textbf{No. of segments}} \\
\cline{3-6}
              &                & \textbf{en$\rightarrow$cs} & \textbf{en$\rightarrow$ja} & \textbf{en$\rightarrow$ru} & \textbf{en$\rightarrow$zh} \\
\hline
\multirow{3}{*}{DA} 
    & train      & 345K & 119K & 350K & 533K \\
    & dev        & 43K  & 15K  & 43K  & 66K  \\
    & test       & 38K  & 13K  & 39K  & 59K  \\
\hline
\multirow{3}{*}{MQM} 
    & train      & --   & --   & 172K & --   \\
    & dev        & --   & --   & 21K  & --   \\
    & test       & --   & --   & 19K  & --   \\
\hline
\multirow{3}{*}{SQM} 
    & train      & 64K  & 75K  & 64K  & 75K  \\
    & dev        & 8K   & 9K   & 8K   & 9K   \\
    & test       & 7K   & 8K   & 7K   & 8K   \\
\hline
\textbf{Total} &            & \textbf{505K} & \textbf{238K} & \textbf{723K} & \textbf{750K} \\
\hline
\end{tabular}
\caption{Dataset statistics by evaluation type, split, and language pair (K represents values in thousands).}
\label{data_stat}
\end{table}

\begin{table*}[ht]
\small
\renewcommand{\arraystretch}{1.4}
    \centering
    \small
    \begin{tabular}{c|cccc|cccc|c}\hline
         &  \multicolumn{4}{c|}{DA}&  \multicolumn{4}{c|}{SQM}& MQM\\
         &  \textbf{en$\rightarrow$ru} & \textbf{en$\rightarrow$cs} & \textbf{en$\rightarrow$ja} & \textbf{en$\rightarrow$zh} & \textbf{en$\rightarrow$ru} & \textbf{en$\rightarrow$cs} & \textbf{en$\rightarrow$ja} & \textbf{en$\rightarrow$zh}& \textbf{en$\rightarrow$ru}\\\hline

\textsc{BertScore}  & 0.399 & 0.480 & 0.418 & 0.328 & 0.290 & 0.207 & 0.290 & 0.107 & -0.04 \\\hline
\textsc{Comet-22}   & 0.571 & 0.637 & 0.511 & 0.443 & 0.450 & 0.400 & 0.352 & 0.220 & 0.075 \\\hline 
\textsc{Comet-22-LS} & 0.848 & \textbf{0.894} & 0.777 & \textbf{0.772} & \textbf{0.572} & 0.701 & \textbf{0.668} & 0.585 & 0.866 \\\hline 
\textsc{Roberta-LS}  & \textbf{0.874} & 0.890 & \textbf{0.780} & 0.770 & 0.557 & \textbf{0.707} & 0.666 & \textbf{0.600} & \textbf{0.874}\\
\hline
\end{tabular}
\caption{System-level Pearson correlation results for MQM, SQM, and DA annotations. Bold values indicate systems that achieved higher correlations with human judgments. LS denotes models trained on long-span input data.}
\label{tab:results_set}
    
\end{table*}

\begin{table*}[t]
\small
\renewcommand{\arraystretch}{1.5}
    \centering
    \begin{tabular}{c|cccll}\hline
          & \textbf{en$\rightarrow$ru} & \textbf{en$\rightarrow$cs} & \textbf{en$\rightarrow$ja} & \textbf{en$\rightarrow$zh} & avg.\\
         \hline
         \textsc{BertScore}&  0.216&  0.344&  0.354& 0.217&0.283\\\hline
 \textsc{Comet-22} & 0.365& 0.519& 0.432&0.331&0.412\\\hline
 
 \textsc{Comet-22-LS}& 0.762& 0.798& 0.722& 0.679&0.740\\ \hline
 \textsc{Roberta-LS}& \textbf{0.768}& \textbf{0.799}& \textbf{0.723}&\textbf{0.685}& \textbf{0.744}\\ \hline
    \end{tabular}
\caption{Segment-level Pearson correlation scores for the language pairs \texttt{en-ru}, \texttt{en-cs}, \texttt{en-ja}, and \texttt{en-zh}. Bold values indicate stronger correlations with human judgments. LS denotes models trained on long-span input data.}

    \label{tab:results_lang}
\end{table*}

\section{Experimental Setup}

Our system is built on top of the COMET package, utilizing the \texttt{comet-train} and \texttt{comet-score} commands to train and evaluate our models. We fine-tuned the pre-trained model \texttt{Unbabel/wmt22-comet-da}, originally trained on DA data, as well as the multilingual pre-trained model \texttt{FacebookAI/xlm-roberta-base} \citep{liu2019roberta}, using A100 GPU. The \texttt{xlm-roberta} model was pre-trained on 2.5TB of filtered CommonCrawl data containing 100 languages and has approximately 279 million parameters.

We trained the models for up to 5 epochs and employed early stopping when the Spearman correlation on the development set did not improve for two consecutive evaluations. For each language pair, the augmented dataset contained five times the original data; however, due to limited GPU memory, we faced out-of-memory issues and restricted training to augmentations with up to two segments. The training process with the augmented dataset took approximately 10 hours for each model.

We trained one baseline model (the fine-tuned version of \texttt{wmt22-comet-da}) and three main models (one primary and two secondary submissions) using the data augmentation approach, retaining only the last two checkpoints for each. Spearman correlation on the test split was used to select the best checkpoint per language pair, and this best model was used for the final submission.

For official WMT \texttt{test25}\footnote{https://github.com/wmt-conference/wmt25-mteval/blob/main/data/testset/mteval-task1-test25.tsv.gz} evaluation, the full segment was used since it fell within the maximum sequence length (512) supported by both the baseline and fine-tuned models. COMET scores typically range between 0 and 1, but can sometimes exceed 1, indicating exceptionally high-quality segments. To align with the ESA metric’s scoring strategy \cite{kocmi2024error}, we upscaled and rounded the scores to a range between 0 and 100.
 
\section{Results}
As described in Section \ref{corpora}, our experiments use the normalized versions of multiple human annotations collected from previous WMT shared tasks. To evaluate and compare our approach, we applied a similar augmentation method to construct long-sequence test data, incorporating annotations from DA, MQM, and SQM. Our baseline sentence-level quality estimation models are \texttt{wmt22-comet-da} (referred to as \textsc{Comet-22}) and \textsc{BertScore}. \textsc{Comet-22-LS}, a fine-tuned version of \texttt{wmt22-comet-da} on long-span data, and \textsc{Roberta-LS}, fine-tuned from \texttt{FacebookAI/xlm-roberta-base}, both serve as long-sequence quality score prediction models. Following \citet{rei-etal-2020-comet}, the models were trained on triplets of (source, hypothesis, reference) and output a score between 0 and 1 reflecting the translation quality relative to both source and reference.

We used the Pearson correlation coefficient to evaluate the models’ performance. Segment-level Pearson correlations on the self-test set are presented in Tables \ref{tab:results_set} and \ref{tab:results_lang}. Our results indicate that metrics from models trained on long-context inputs generally outperform sentence-level metrics, in some cases by a significant margin. 


The annotation-wise segment-level correlation results in Table \ref{tab:results_set} demonstrate that the unsupervised baseline metric, \textsc{BertScore}, exhibits relatively weak correlations across all language pairs. The sentence-level \textsc{Comet-22} model shows improved correlations, especially for DA annotations, reflecting its training on DA data. Moreover, it outperforms \textsc{BertScore} on SQM and MQM annotations, indicating its ability to mimic human annotations by learning from data. Our fine-tuned long-sequence baseline model, \textsc{Comet-22-ls}, surpasses the sentence-level baselines, achieving performance close to, and in some cases better than, our primary submission models based on \textsc{Roberta-ls}. Notably, \textsc{Comet-22-ls} achieves results on MQM annotations that are very close to those of \textsc{Roberta-ls}, outperforming sentence-level baselines by a substantial margin. This suggests that training on longer context sequences provides considerable benefits, especially for complex annotation types like MQM, which capture fine-grained translation errors.

Table \ref{tab:results_lang} summarizes correlation results across all language pairs using joint annotations. Across nearly all language pairs, our models outperform baseline metrics in correlation with human judgments. These results suggest that combining multiple segments within the same document or domain is more effective than independently scoring segmented sentences and averaging their scores \citep{raunak2024slidereferencefreeevaluationmachine}. The improvement may be attributed to the model’s ability to capture contextual information across long-sequence segments, thereby enabling more context-aware quality estimation.

\begin{figure}
    \centering
    \includegraphics[width=1\linewidth]{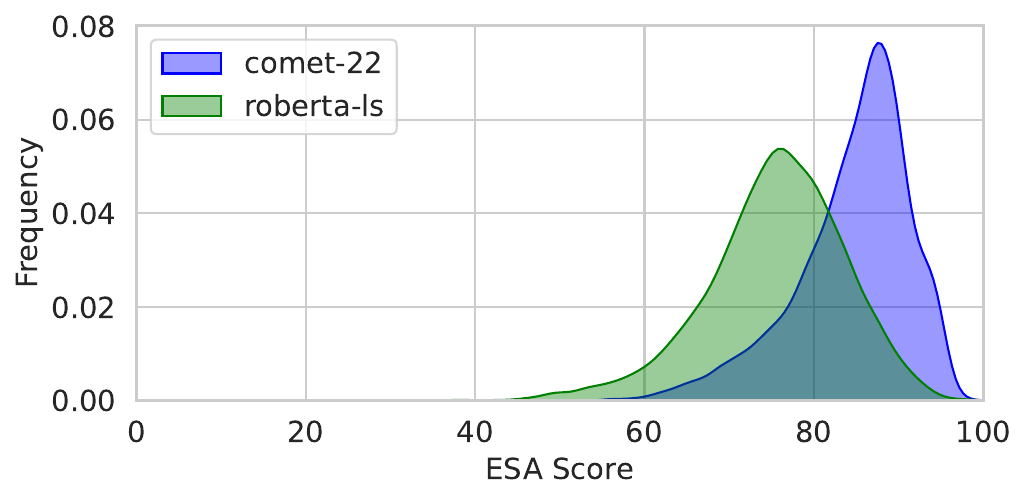}
    \caption{Distribution of segment-level scores assigned by sentence-level \textsc{Comet-22} and long-sequence \textsc{Roberta-ls} model.}
    \label{fig:score_distribution}
\end{figure}

However, handling longer text sequences poses challenges due to the input size limitations of the underlying models, as highlighted by \citet{gong2020recurrentchunkingmechanismslongtext}. This necessitates careful segmentation and score-averaging strategies to compute scores at the paragraph or document level. We also conduct a preliminary analysis of the score distributions (Figure~\ref{fig:score_distribution}) and find that sentence-level baseline scores (\textsc{Comet-22}) are mostly concentrated between 60 and 100, with a pronounced peak around 90. In contrast, the \textsc{Roberta-ls} model, trained on multi-sentence inputs, produces a more widely spread score distribution that better reflects the variability typically observed in human judgments \citep{toral_attaining_2018}. This wider spread may be due to the long-span training data being based on weighted average scores that encompass a broader range of score scales. By contrast, the narrower distribution of \textsc{Comet-22} scores could stem from the characteristics of the human-annotated data on which it was trained, where non-expert evaluators have been shown to assign disproportionately higher fluency and adequacy ratings, resulting in smaller score gaps and reduced variance compared to expert assessments \citep{toral_attaining_2018}. This indicates that models trained with longer context are more sensitive to subtle quality differences, reflecting a more nuanced understanding of translation quality.

\section{Related Work}
In recent years, metrics based on large pre-trained models have emerged as strong alternatives to traditional n-gram-based approaches, enabling better capture of semantic similarity between words beyond mere lexical matching. These metrics broadly fall into two categories: embedding-based metrics and fine-tuned metrics.

Embedding-based metrics typically represent an advancement over n-gram matching by using dense word representations in an embedding space to compute scores that reflect semantic similarity between reference and hypothesis segments. Notable examples include YISI-1 \citep{lo-2019-yisi}, MOVERSCORE \citep{chow-etal-2019-wmdo}, and BERTSCORE \citep{zhang2019bertscore}, which leverage embedding models for soft alignment between two segments to capture semantic similarity effectively.

Fine-tuned metrics, on the other hand, involve learnable models such as RUSE \citep{shimanaka-etal-2018-ruse}, BLEURT \citep{sellam-etal-2020-bleurt}, and COMET \citep{rei-etal-2020-comet, rei-etal-2022-comet} that directly optimize underlying embedding models to maximize correlation with human judgments. These models have demonstrated promising results in producing reliable quality scores for test sets such as DA or MQM. While most of these metrics perform reference-based evaluation, recent advancements leveraging highly multilingual pre-trained encoders like multilingual BERT \citep{devlin2018bert} and RoBERTa \citep{liu2019roberta, conneau2019unsupervised} have enabled reference-less systems to show encouraging correlations with human judgments \citep{freitag2023results}.

Most automatic evaluation approaches rely on decontextualized assessments, where translations are judged at the sentence level. However, sentences are often inherently ambiguous, and incorporating document-level context has been shown to be beneficial for both MT and its evaluation \citep{laubli-etal-2018-machine, castilho-etal-2020-context, castilho_survey_2024, vernikos2022embarrassingly}. Consequently, a few automatic metrics have been developed to extend evaluation beyond the word or sentence level \citep{vernikos2022embarrassingly, jiang2021blonde}. These metrics aim to address discourse-level phenomena such as lexical consistency, coherence, ellipsis, and pronoun resolution \citep{voita2018context, bawden2017evaluating}.

However, existing methods typically use a limited number of surrounding sentences as context, allowing models to incorporate neighboring information when embedding each sentence and computing scores at the sentence level \citep{rei2020unbabel, vernikos2022embarrassingly, hu2023exploring}. In contrast, long-sequence or document-level evaluation processes the entire segment as a single input, enabling deeper discourse-level resolution and offering a promising yet still underexplored avenue for improving alignment with human judgments.
\section{Conclusion}

In this paper, we present DCU\_ADAPT’s contribution to the WMT25 MT Evaluation Shared Task. We leverage the COMET framework and train regression models to predict ESA quality scores. In line with the Shared Task goals, we augment the provided training data and optimize our models to evaluate long, multi-sentence units of text. By fine-tuning multilingual models for cross-lingual transfer, we utilize source, reference, and hypothesis as inputs. Our primary submission — a fine-tuned pre-trained model trained on augmented data — demonstrates higher or otherwise competitive correlation levels with human judgments across multiple languages. Further investigation comparing long-text evaluation after segmentation with sentence-level evaluation is a promising direction for future work.

The data and code produced during this Shared Task participation are available at: \url{https://github.com/sami-haq99/CAEMT/tree/main/wmt-2025-submission}.

\section*{Limitations}
We trained our models using augmented long-segment level scores from the MQM, DA, and SQM datasets. However, we only evaluated the models on self-test data carefully extracted from the training set; evaluating on benchmark datasets will better clarify the true benefits of our approach. Additionally, we normalized the data using min-max normalization to combine different datasets and upscaled the predicted scores to match the ESA metric’s score range. Furthermore, our training and testing were conducted on GPUs with at least 40GB of memory; due to time constraints, we were unable to evaluate performance on CPUs. 

\section*{Ethics Statement}
Our research focuses on evaluating long-sequence outputs of MT systems using quality estimation models trained on augmented data. We are committed to conducting and reporting our evaluations with the highest levels of transparency and fairness. By upholding these principles, we aim to contribute to reliable and objective assessment practices in MT evaluation.

\section*{Acknowledgements}
This work was conducted with the financial support of the Research Ireland Centre for Research Training in Digitally-Enhanced Reality (d-real) under Grant No. 18/CRT/6224, and Research Ireland Centre for Research Training in Artificial Intelligence under Grant No. 18/CRT/6223. For the purpose of Open Access, the author has applied a CC BY public copyright licence to any Author Accepted Manuscript version arising from this submission.

The Authors also benefit from being members of the ADAPT SFI Research Centre at Dublin City University, funded by the Science Foundation Ireland under Grant Agreement No. 13/RC/2106\_P2.

\bibliography{emnlp2023-latex/references}
\bibliographystyle{acl_natbib}

\end{document}